# DUAL-SPACE KNOWLEDGE DISTILLATION WITH KEY-QUERY MATCHING FOR LARGE LANGUAGE MODELS WITH VOCABULARY MISMATCH

*Stella Eva Tsiapali*[⋆] *Cong-Thanh Do*[†] *Kate Knill*[⋆]

[⋆] University of Cambridge, Department of Engineering, Cambridge, United Kingdom
[†] Toshiba Europe Limited, Cambridge, United Kingdom

## ABSTRACT

Large language models (LLMs) achieve state-of-the-art (SOTA) performance across language tasks, but are costly to deploy due to their size and resource demands. Knowledge Distillation (KD) addresses this by training smaller Student models to mimic larger Teacher models, improving efficiency without substantial performance loss. Dual-Space Knowledge Distillation with Cross-Model Attention (DSKD-CMA) has emerged as a SOTA method for KD between LLMs with distinct tokenizers, yet its internal workings remain largely opaque. In this work, we systematically analyse the attention mechanism of DSKD-CMA through manual token alignment probing and heatmap visualisations, revealing both strengths and limitations. Building on this, we introduce a novel method, DSKD-CMA-GA, based on Generative Adversarial (GA) learning, to address the mismatched distributions between the keys and queries computed from distinct models. Experiments show modest but consistent ROUGE-L gains in text generation quality, particularly on out-of-distribution data (+0.37 on average), narrowing the gap between cross- and same-tokenizer KD[1].

## 1. INTRODUCTION

The rise of large language models (LLMs) has driven major advances in text generation and reasoning, yet their scale makes deployment costly in computation, latency, and energy [1]. Knowledge Distillation (KD) mitigates this by transferring capabilities from large Teacher models to smaller Student models, preserving performance while improving efficiency.

Meanwhile, current LLMs are built with distinct tokenizers and vocabularies, complicating the comparison and alignment of their outputs [2]. Recent works have proposed approaches to overcome this challenge, among which Dual-Space Knowledge Distillation with Cross-Model Attention (DSKD-CMA) [3] currently represents a state-of-the-art (SOTA) method. Yet, despite its empirical success, its internal workings remain poorly understood, hindering development.

This paper pursues two goals: (1) to analyse the behaviour and limitations of DSKD-CMA, and (2) propose an enhanced framework that improves cross-tokenizer KD. Hence, our contributions are:

1. **Analysis framework for DSKD-CMA:** We introduce multi-token chunk alignment probes and heatmap visualisations of intermediate attention states, revealing both strengths (chunk-oriented alignment) and weaknesses (limited localisation).
2. **Key-query matching distillation objectives:** We propose DSKD-CMA-GA and DSKD-CMA-CT, which extend DSKD-CMA with key-query matching, based on Generative Adversarial (GA) learning and Conditional Transport (CT), respectively. These improve text generation quality across benchmarks and reach, or surpass, same-tokenizer KD performance.

[1] Our code is publicly available at: `https://github.com/stellaevat/DSKD-KQ`.

## 2. BACKGROUND AND RELATED WORK

### 2.1. Knowledge Distillation (KD)

KD transfers knowledge from a large Teacher model to a smaller Student model, improving efficiency with minimal performance loss. Instead of relearning from data, the Student mimics the Teacher's behaviour to capture its emergent abilities [1].

In its simplest form, black-box KD, the Student is trained directly on Teacher outputs [4, 5]. While memory-efficient, this largely reduces to next-token prediction and limits knowledge transfer. More sophisticated white-box KD methods instead leverage internal states, aligning logits [6, 7] or intermediate representations [8, 9], sometimes employing Reinforcement Learning (RL) principles for on-policy training [10]. Logit-based KD offers a balance between transfer and GPU memory demands [1], but assumes Teacher and Student token sequences of equal length and shared vocabulary dimensions, which is often not feasible across LLMs.

### 2.2. Cross-Tokenizer KD

Current LLMs employ distinct tokenizers with different vocabularies, producing sequences of varying lengths and predictive spaces. Thus, both sequence and vocabulary alignment need to be addressed, in order to be able to compare their internal or output states.

For sequence alignment, some methods identify one-to-one token mappings via minimum edit-distance (MinED) [11, 12], though these are sensitive to small mismatches. Others focus on multi-token mappings, grouping tokens that span the same text into chunks, and aggregating their hidden states (e.g., mean, product) [2, 13]. This improves alignment fidelity, but risks information loss due to the aggregation. Subsequently, vocabulary alignment approaches include edit-distance mappings [12], top-K logit truncation [13], or more principled methods like Universal Logit Distillation (ULD), which applies Optimal Transport (OT) between the Teacher and Student distributions [14], although at a higher computational cost.

By contrast, Dual-Space Knowledge Distillation with Cross-Model Attention (DSKD-CMA) [3] addresses sequence and vocabulary alignment jointly. It employs projectors and attention to efficiently learn implicit alignments, achieving SOTA results. Yet, unlike explicit alignment methods, its inner workings are less interpretable. $\text{COT}_2\text{ALIGN}$ [15] improves performance by adding

OT-based Chain-of-Thought (CoT) alignment to DSKD-CMA, but leads to a much higher computational cost. Hence, the in-depth analysis of DSKD-CMA may lead both to better understanding and more informed extensions, which directly address its internal issues.

## 3. METHODS

### 3.1. Original DSKD-CMA Method

To demonstrate how our methods fit into the existing framework, we provide a brief overview of DSKD-CMA [3].

A **Cross-Model Attention** (CMA) mechanism is employed to align Teacher and Student hidden states of distinct dimensions. The Student embeddings are projected into the Teacher space to form queries, $Q$, while the Teacher embeddings serve as keys, $K$. Hence, for a Student sequence with (length × hidden dimension) = $(n_s \times d_s)$, and a Teacher sequence $(n_t \times d_t)$:

$$Q = P_Q([\mathbf{e}^s_{\text{input}}, \mathbf{e}^s_{\text{target}}]) \in \mathbb{R}^{n_s \times 2d_t}, \tag{1}$$

$$K = N_{std.}([\mathbf{e}^t_{\text{input}}, \mathbf{e}^t_{\text{target}}]) \in \mathbb{R}^{n_t \times 2d_t}, \tag{2}$$

where $P_Q$ is a learnable projector and $N_{std.}$ denotes standard deviation normalisation for faster convergence.

Attention matrices are then computed via scaled and softmaxed dot products:

$$A^{t\to s} = \sigma\left(\frac{QK^T}{\sqrt{2d_t}}\right) \in \mathbb{R}^{n_s \times n_t},\ A^{s\to t} = \sigma\left(\frac{KQ^T}{\sqrt{2d_t}}\right) \in \mathbb{R}^{n_t \times n_s}. \tag{3}$$

Thus, the Student and Teacher hidden states, $\mathbf{h}^s$ and $\mathbf{h}^t$, are projected to each other's hidden space using learnable projectors:

$$V^{s\to t} = P^{s\to t}(\mathbf{h}^s) \in \mathbb{R}^{n_s \times d_t}, \tag{4}$$

$$V^{t\to s} = P^{t\to s}(N_{std.}(\mathbf{h}^t) + N_{std.}(\mathbf{e}^t_{\text{target}})) \in \mathbb{R}^{n_t \times d_s}, \tag{5}$$

and to each other's sequence length, using the attention matrices:

$$\mathbf{h}^{t\to s} = A^{t\to s} V^{t\to s}, \quad \mathbf{h}^{s\to t} = A^{s\to t} V^{s\to t}. \tag{6}$$

Figure 1 illustrates this process in terms of the sequence of matrix multiplications involved.

The projected states are then passed through the opposite model's prediction head, $W^s$ or $W^t$, to yield output distributions:

$$q^{t\to s} = \sigma(\mathbf{h}^{t\to s} W^s), \quad q^{s\to t} = \sigma(\mathbf{h}^{s\to t} W^t), \tag{7}$$

in order to compute the loss in that space under some $f$-divergence:

$$L^{t\to s}_{KD} = \sum_{i \le n_s} f(\boldsymbol{p}^s || \boldsymbol{q}^{t\to s}), \quad L^{s\to t}_{KD} = \sum_{j \le n_t} f(\boldsymbol{p}^t || \boldsymbol{q}^{s\to t}). \tag{8}$$

These are jointly minimised, along with the cross-entropy of the Student space distributions ($p^s$ and $q^{t\to s}$) against gold targets:

$$L_{DSKD-CMA} = \tfrac{1}{2} L^s_{CE} + \tfrac{1}{2}(L^{s\to t}_{KD} + L^{t\to s}_{KD} + L^{t\to s}_{CE}). \tag{9}$$

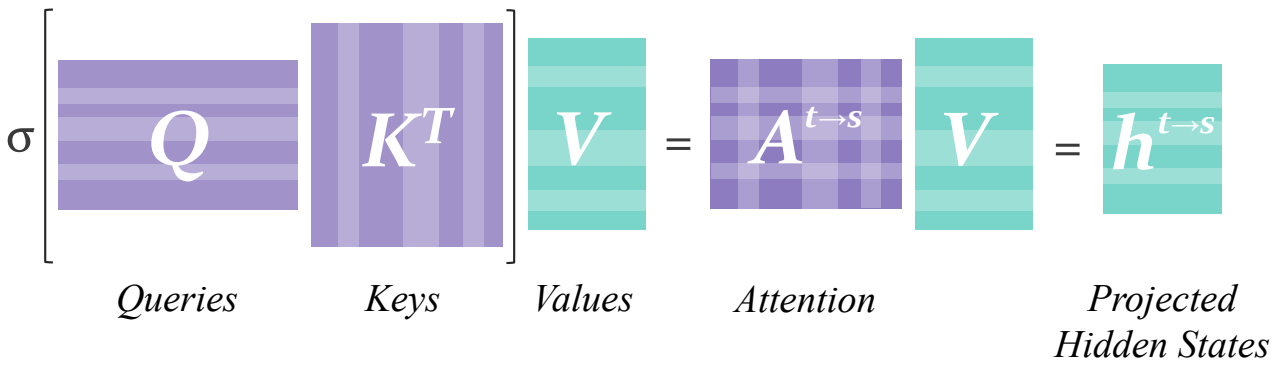

**Fig. 1**: The sequence of matrix multiplications employed by the attention mechanism of DSKD-CMA.

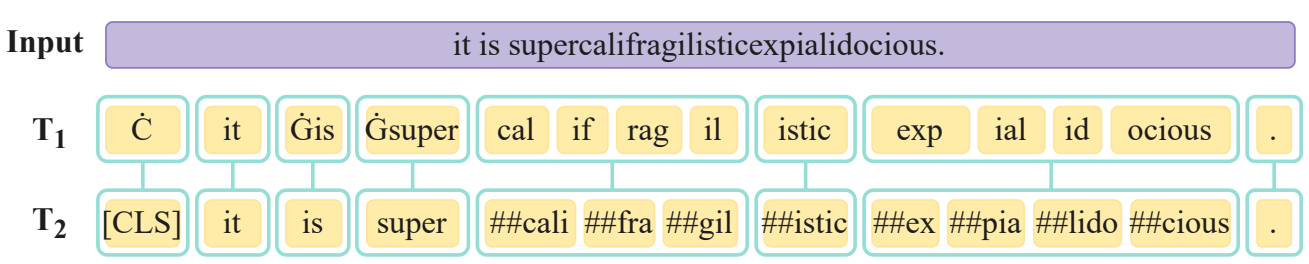

**Fig. 2**: Chunk alignment of token sequences by distinct tokenizers.

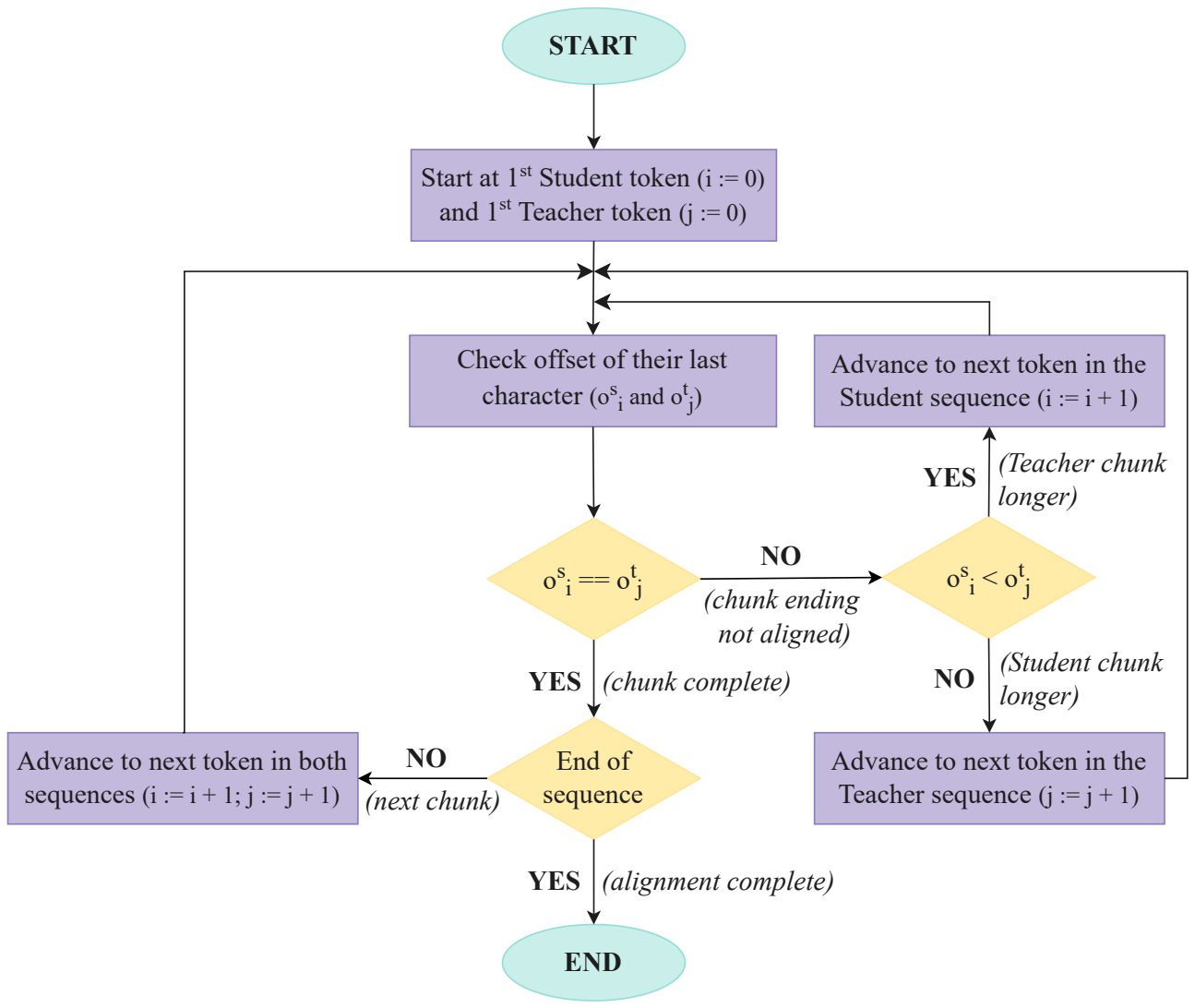

**Fig. 3**: Flowchart of the proposed manual chunk alignment process.

### 3.2. Proposed Cross-Model Attention Analysis

While effective, CMA is opaque: it implicitly learns alignments, offering no insights into how tokens influence each other. To probe this, we construct explicit **chunk-based alignments**, segmenting Teacher and Student tokenizations into minimal chunks that span the same text, as illustrated in Figure 2.

Similar to [2], for Teacher and Student tokenization and decoding functions $T_t/D_t$ and $T_s/D_s$, respectively, the start- and end-token indices of the aligned chunks in input text $\mathbf{x}$ are given by:

$$\begin{aligned} C(\mathbf{x}) = \{(i,j,k,l) \in \mathbb{Z}^4 \mid\ & D_t(T_t(\mathbf{x})_{:i}) = D_s(T_s(\mathbf{x})_{:k}), \\ & D_t(T_t(\mathbf{x})_{i:j}) = D_s(T_s(\mathbf{x})_{k:l})\} \end{aligned} \tag{10}$$

In our implementation, character offsets are generated for each token during tokenization, and processed incrementally during alignment. As shown in Figure 3, we compare the last-character offsets of each Teacher and Student token, advancing by one token in whichever sequence has the earlier offset, until they match. When a match occurs, the aligned chunks are recorded, and the next chunk begins from the following token. Using offsets instead of tokens avoids repeatedly decoding subsequences at different lengths.

These token indices are then used to generate chunk alignment matrices, $M^{t\to s} \in \mathbb{R}^{n_s \times n_t}$ and $M^{s\to t} = (M^{t\to s})^T \in \mathbb{R}^{n_t \times n_s}$, between a Teacher sequence of length $n_t$ and a Student sequence of length $n_s$, where only the index ranges corresponding to aligned chunks contain non-zero values. Thus, for $(i,j,k,l) \in C(\mathbf{x})$:

$$M^{t\to s}_{i:j,k:l} = 1,\ M^{t\to s}_{i:j,:k} = 0,\ M^{t\to s}_{i:j,l:} = 0. \tag{11}$$

This is an interpretable mapping against which CMA's learned alignments can be compared, both in terms of performance and heatmap visualisations. Thus, we introduce two ablations of CMA:

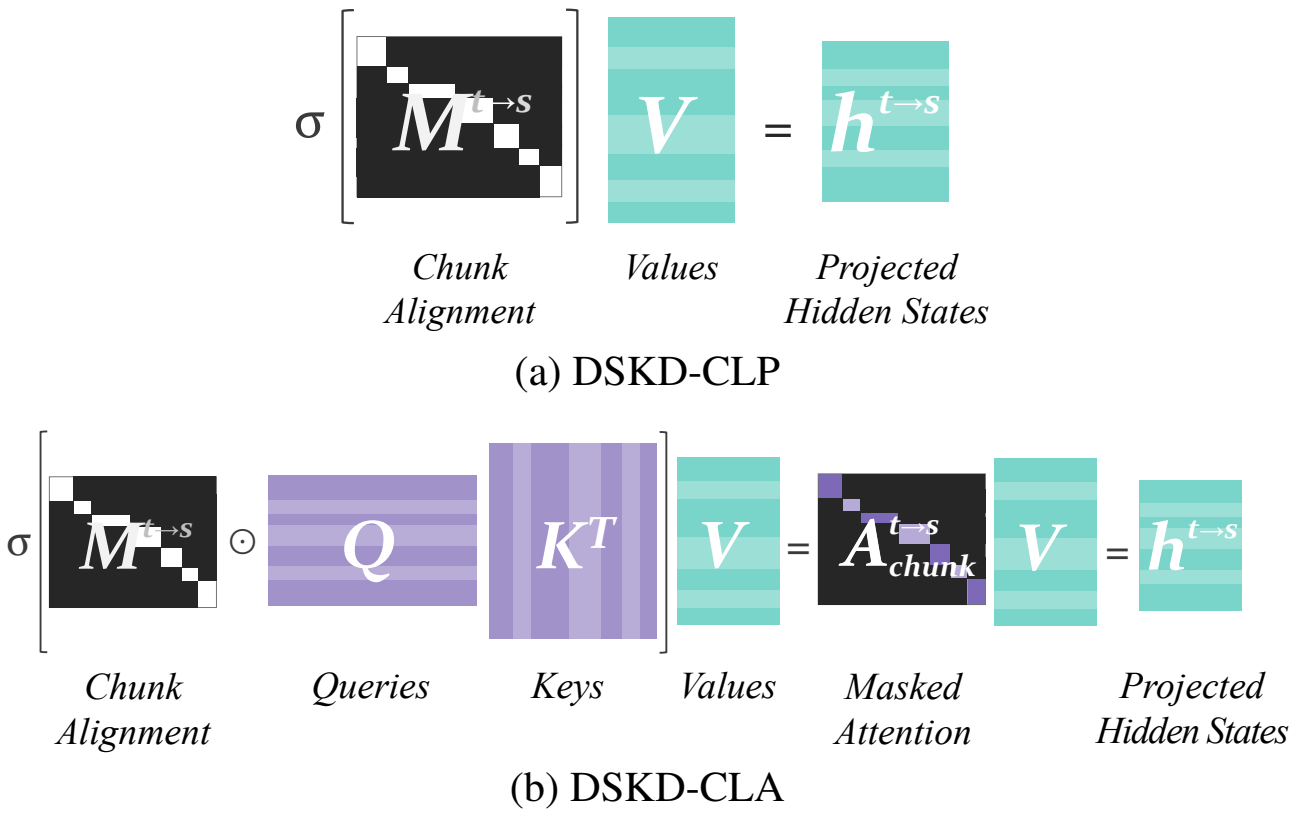

(a) DSKD-CLP

(b) DSKD-CLA

**Fig. 4**: The sequence of matrix multiplications involved in Chunk-Level Projection (CLP) and Chunk-Level Attention (CLA).

**Chunk-Level Projection (CLP).** Here, CMA is replaced by a uniform softmaxed projection between the tokens of to aligned chunks:

$$\mathbf{h}^{t\to s} = \sigma(M^{t\to s})V^{t\to s}, \tag{12}$$

and respectively for $\mathbf{h}^{s\to t}$. This tests whether simple chunk-level projection suffices, without the need for attention.

**Chunk-Level Attention (CLA).** Instead of replacing CMA, we mask it so attention only applies within aligned chunks:

$$A^{t\to s}_{\text{chunk}} = \sigma\left(M^{t\to s} \odot \frac{QK^T}{\sqrt{2d_t}}\right), \tag{13}$$

and respectively for $A^{s\to t}_{\text{chunk}}$. When replacing the original attention mechanism from Equation 4, CLA retains adaptive weighting between tokens based on the similarity of their embeddings, $Q$ and $K$, but it enforces a more interpretable, localised chunk-based alignment. Figure 4 illustrates these processes in terms of the sequence of matrix multiplications involved.

### 3.3. Proposed Key-Query Matching Distillation

The CMA mechanism is computed from Teacher and Student embeddings, used as *keys* and *queries*. Since these stem from distinct models, they may have mismatched distributions, $p_K$ and $p_Q$, weakening the effectiveness of the attention mechanism. To address this, we propose to integrate **key-query matching** into DSKD-CMA. We experiment with two variants, adapted from [16]:

**Generative Adversarial Alignment (GA).** A discriminator, $D$, learns to distinguish Teacher keys from Student queries, while the Student's query projector, $P_Q$, is the generator that learns to fool it:

$$L_{\text{KQ}} = \min_{P_Q} \max_{D} \left( \mathop{\mathbb{E}}_{q\sim p_Q}[\log D(q)] + \mathop{\mathbb{E}}_{k\sim p_K}[\log(1 - D(k))] \right). \tag{14}$$

**Conditional Transport Alignment (CT).** Using a constrained version of Optimal Transport [17] to reduce computational load, bidirectional matching between key and query distributions is encouraged via a trainable critic, $c$. This also acts adversarially, maximising the overall transport cost, while $P_Q$ tries to minimise it:

$$L_{\text{KQ}} = \min_{P_Q} \max_{c} \tfrac{1}{2}\mathbb{E}[c(k,q)] + \tfrac{1}{2}\mathbb{E}[c(k,q)]. \tag{15}$$

In both cases, the overall distillation objective proposed is:

$$L_{\text{train}} = L_{\text{DSKD-CMA}} + L_{\text{KQ}}. \tag{16}$$

## 4. EXPERIMENTAL SETUP

**Datasets** Following [3], we use the DataBricks Dolly 15K dataset [18] for distillation, with a 11K-1K train-validation split. For evaluation, we test in-distribution on **Dolly** (500 samples) and out-of-distribution on Self-Instruct (**SelfInst**, 242 samples) [19], Vicuna-Eval (**Vicuna**, 80 samples) [20], Super-Natural Instructions (**S-NI**, 1,649 samples) [21], and Unnatural Instructions (**UnNI**, 23,916 samples) [22], totalling 26,387 test samples. All datasets have been preprocessed for KD by [23].

**Models** The Student model used is **GPT-2** (124M params, 50K vocab). We use Teacher **GPT-2 XL** (1.5B params, 50K vocab) for same-tokenizer KD, and **Qwen1.5** (1.8B params, 150K vocab) for cross-tokenizer KD. The large vocabulary size mismatch between Qwen1.5 and GPT-2 makes this setting particularly challenging and suitable for evaluation.

**Baselines** We compare against the original DSKD-CMA method, as well as two prominent cross-tokenizer KD baselines: **MinED** [12] and **ULD** [14]. Additionally, we compare with non-distilled supervised fine-tuning on the same Dolly training data (**SFT**), and the typically more successful same-tokenizer DSKD variant [3], in order to assess their performance gap.

**Divergence Measures** We test six established divergence functions for Equation 8, which capture a range of mode-covering vs. mode-seeking behaviours and allow comprehensive comparison to [3]: (a) Kullback-Leibler (**KL**); (b) reverse KL (**RKL**) [24]; (c) skewed KL (**SKL**); (d) skewed reverse KL (**SRKL**) [25]; (e) adaptive KL (**AKL**) [26]; and (f) Jensen-Shannon divergence (**JSD**) [27]. We report results obtained with the best divergence function for each method.

**Metric** Following prior work, we report **ROUGE-L** scores [28], which capture the longest common subsequence overlap between the generated and reference text. This metric balances precision with recall, rewarding both completeness and relevance.

**Hardware** Training and evaluation were run on an Ampere node with 4 NVIDIA A100 GPUs (80GB each) and 1TB RAM. Each training session required 2-3 hours, while evaluation lasted approximately 1 hour per variant.

## 5. RESULTS AND DISCUSSION

Table 1 summarises results for the Student and Teacher models, KD baselines and all DSKD variants tested. The initial Student achieves just 20-50% of the Teachers' performance, highlighting the gap that KD has to close.

### 5.1. Chunk-Based Probing Insights

To better understand the role of CMA, we compared it against chunk-based alternatives. Replacing CMA with Chunk-Level Projection (**CLP**), where Teacher and Student hidden states are explicitly projected onto matched chunks, yields similar performance, and in some cases slightly outperforms CMA. Figure 5a demonstrates that the alignments produced by CMA indeed tend to capture chunk-level structure as intended, but can be noisy in the presence of repeated (e.g., numerical) tokens. This issue is eliminated by the strictly localised alignment of CLP. However, CLP is more computationally expensive, requiring explicit alignment at every step, which makes CMA a more practical solution in large-scale training.

| | ID | OOD | | | | Avg. | |
|---|---|---|---|---|---|---|---|
| | **Dolly** | **SelfInt** | **Vicuna** | **S-NI** | **UnNI** | **OOD** | **All** |
| Student (GPT-2) | 6.39 | 4.72 | 10.18 | 4.49 | 4.64 | 6.01 | 6.08 |
| **Same-Tokenizer KD** | | | | | | | |
| Teacher (GPT-2 XL) | 27.07 | 14.65 | 16.24 | 27.12 | 31.41 | 22.36 | 23.30 |
| DSKD [3] | 25.98 | 10.58 | 15.54 | 19.22 | 22.13 | 16.87 | 18.69 |
| **Cross-Tokenizer KD** (with best $f$-divergence, where applicable) | | | | | | | |
| Teacher (Qwen1.5) | 27.45 | 19.42 | 20.93 | 34.74 | 35.98 | 27.77 | 28.30 |
| SFT | 22.83 | 9.49 | 14.66 | 13.55 | 16.06 | 13.44 | 15.32 |
| MinED [12] | 21.69 | 9.78 | 15.59 | 14.98 | 17.25 | 14.40 | 15.86 |
| ULD [14] | 21.73 | 9.28 | 14.85 | 15.17 | 17.15 | 14.11 | 15.64 |
| DSKD-CMA [3] | 24.64 | 11.50 | **16.21** | 18.40 | 21.69 | 16.95 | 18.49 |
| DSKD-CLP | 24.52 | 11.35 | 15.28 | 19.26 | **22.12** | 17.00 | 18.51 |
| DSKD-CLA | 24.20 | 11.44 | 14.93 | 18.53 | 21.32 | 16.55 | 18.08 |
| DSKD-CMA-GA | 24.79 | **11.82** | 15.95 | **19.44** | 22.05 | **17.32** | **18.81** |
| DSKD-CMA-CT | 24.63 | 10.83 | 16.08 | 17.87 | 21.07 | 16.46 | 18.10 |
| **Cross-Tokenizer KD** (with KL divergence) | | | | | | | |
| DSKD-CMA [3] | 24.13 | 9.73 | 15.20 | 15.14 | 17.78 | 14.46 | 16.40 |
| DSKD-CMA-GA | **24.92** | 11.17 | 15.84 | 18.19 | 21.50 | 16.68 | 18.32 |
| DSKD-CMA-CT | 24.63 | 10.83 | 16.08 | 17.87 | 21.07 | 16.46 | 18.10 |

**Table 1**: Mean ROUGE-L scores from evaluation on five random seeds. Average scores across the out-of-distribution (**OOD**) and all (**All**) evaluation datasets are given. Results for the best $f$-divergence are reported for DSKD-CMA (SRKL), -GA (SKL) and -CT (KL).

We also experimented with Chunk-Level Attention (**CLA**), which applies the attention mechanism only within each aligned chunk, to refine and localise mappings. This variant degrades performance on most datasets. Figure 5c reveals that this fine-grained weighting emphasises within-chunk token relationships, weighting individual tokens much more heavily than others. Therefore, DSKD seems to benefit from stable coarse-grained alignments, rather than precise but over-sensitive to local relationships mappings.

Notably, all DSKD variants drastically outperform previous work [12, 14], showcasing the effectiveness of projecting the Teacher-Student states to each other's space for comparison, irrespective of the specific implementation.

### 5.2. Key-Query Matching Improvements

The success of DSKD-CMA heavily relies on the effectiveness of its attention mechanism, especially when compared to same-tokenizer KD. An important issue is that the keys and queries used stem from distinct models, and hence distributions, so their direct comparison can be unreliable. To address this, we proposed DSKD-CMA-GA and DSKD-CMA-CT, which use key-query matching [16] to align Teacher and Student representations.

Among the approaches tested, DSKD-CMA-GA achieves the strongest performance. It yields modest but consistent improvements over the DSKD-CMA baseline across both ID and OOD datasets, with ROUGE-L gains of 0.15–1.04 points, depending on the dataset. DSKD-CMA-CT lags behind, indicating that generative adversarial learning may capture alignment under tokenizer mismatch better than conditional transport. The benefits of including GA are more pronounced when comparing methods based on the same (KL) divergence, improving the OOD and overall average by 2.22 and 1.98 points, respectively. Crucially, DSKD-CMA-GA consistently narrows the gap between same- and cross-tokenizer performance, sometimes even surpassing same-tokenizer scores.

It should be noted that $f$-divergence choice interacts with the alignment mechanism employed. By experimenting with all six divergence measures for each method, we have found that SRKL works best for DSKD-CMA (as also reported by [3]), while SKL and KL amplify the benefits of DSKD-CMA-GA and DSKD-CMA-CT, respectively. Thus, divergence function and alignment method should be co-designed for optimal performance.

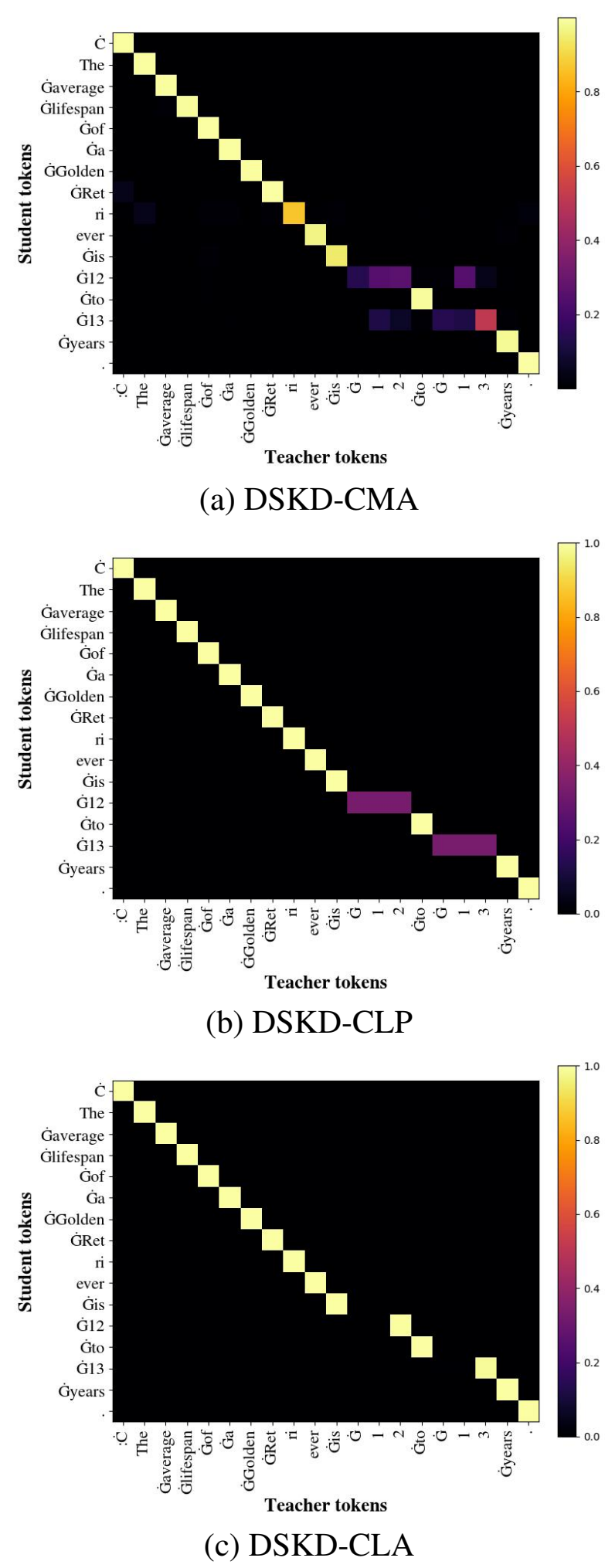

(a) DSKD-CMA

(b) DSKD-CLP

(c) DSKD-CLA

**Fig. 5**: Alignment weights assigned to pairs of tokens in the Student and Teacher sequences, using CMA and its chunk-based ablations: Chunk-Level Projection (CLP) and Chunk-Level Attention (CLA).

## 6. CONCLUSION

This paper has presented a methodical analysis and extension of DSKD-CMA [3], a SOTA method in cross-tokenizer KD. Through chunk-level alignment experiments, we confirmed that CMA implicitly captures the expected chunk structure of token sequences, while also revealing weaknesses in the localisation of its mappings.

Based on this insight, we proposed key-query matching techniques [16] as a principled way to improve cross-tokenizer alignment, exploring both Generative Adversarial (GA) and Conditional Transport (CT) variants. The GA-based approach, combined with the skewed KL divergence, consistently outperformed DSKD-CMA across benchmarks, surpassing even same-tokenizer distillation, which is promising for the potential of cross-tokenizer KD. More challenging reasoning benchmarks [29] and downstream task performance [10] can be tested in future work, for a richer evaluation landscape. Importantly, these results open promising research directions, such as experimentation with dynamic weighting of the proposed loss terms, and the broader application of adversarial learning in the advancement of cross-tokenizer KD.

## 7. ACKNOWLEDGMENTS

No funding was received for conducting this study. The authors have no relevant financial or non-financial interests to disclose.

## 8. COMPLIANCE WITH ETHICAL STANDARDS

This study is on the training and evaluation of machine learning models, so no ethical approval was required.